\documentclass[conference]{IEEEtran}

\usepackage{amsmath,amssymb,amsfonts}
\usepackage{algorithmic}
\usepackage{graphicx}
\usepackage{textcomp}
\usepackage{xcolor}
\usepackage{listings}
\usepackage{hyperref}
\usepackage{caption}
\usepackage[newfloat]{minted}
\def\BibTeX{{\rm B\kern-.05em{\sc i\kern-.025em b}\kern-.08em
    T\kern-.1667em\lower.7ex\hbox{E}\kern-.125emX}}

\newcommand{\RNum}[1]{\uppercase\expandafter{\romannumeral #1\relax}}
\DeclareMathAlphabet{\mathpzc}{OT1}{pzc}{m}{it}

\definecolor{code_gray}{rgb}{.947,.947,.947}

\newenvironment{mycode}{\captionsetup{type=listing}}{}

\begin{document}

\title{GoldenTransformer: A Modular Fault Injection Framework for Transformer Robustness Research\\
}

\author{\IEEEauthorblockN{Luke Howard}
\IEEEauthorblockA{\textit{Radnor High School}, Radnor, USA \\
lhhoward.125@gmail.com}
}

\maketitle

\begin{abstract}
Transformers have become the foundation for a wide range of state-of-the-art models across natural language processing, computer vision, and other machine learning domains. Despite their widespread deployment, the robustness of these models under fault conditions remains underexplored. We present GoldenTransformer, a modular and extensible fault injection framework designed to evaluate the resiliency of Large Language Models (LLMs) to induced hardware faults. GoldenTransformer offers a unified Python-based platform for injecting diverse classes of faults—such as weight corruption, activation injections, and attention-level disruptions—into pretrained transformer-based models. Inspired by the GoldenEye simulator for DNNs, our framework focuses on the unique challenges of working with large transformer architectures, including considerations such as structural complexity, latent dependencies, and nonuniform layer definitions. GoldenTransformer is built atop PyTorch and HuggingFace Transformers, and it supports experiment reproducibility, metric logging, and visualization out of the box. We detail the technical design and use of GoldenTransformer and demonstrate through several example experiments on classification and generation tasks. By enabling controlled injection of faults at multiple logical and structural points in a transformer, GoldenTransformer offers researchers and practitioners a valuable tool for model robustness analysis and for guiding dependable system design in real-world LLM applications.
All code for GoldenTransformer can be found at \href{https://github.com/FuzzyNum/goldentransformer}{https://github.com/FuzzyNum/goldentransformer}
\end{abstract}


\section{Introduction}
The deployment of Large Language Models (LLMs) in real-world systems has brought significant attention to their robustness under abnormal or faulty operating conditions. While these models have demonstrated extraordinary performance across a broad spectrum of natural language processing tasks, their internal complexity and computational requirements render them susceptible to faults at various levels—from low-level weight corruption to high-level architectural failures. Existing fault injection tools, however, are either too general or tailored for conventional neural networks and fail to capture the nuanced structure of transformer-based architectures. To address this gap, we present GoldenTransformer, a modular and extensible fault injection framework specifically designed for transformer-based LLMs.

GoldenTransformer provides an open-source framework for fault injection and performance evaluation specifically targeting LLMs. Its design principles draw from prior work such as GoldenEye~\cite{mahmoud2022goldeneye} but adapt the approach to fit the structural and semantic characteristics of modern LLM architectures. Our contributions include: 

\begin{itemize}
\item An extensible framework for injecting various types of faults—including layer-level perturbations, weight-level corruptions, and activation-level distortions— into transformer-based models.

\item With Built-in metric and visualization support, GoldenTransformer provides documentation and tool for replicable fault experiments across architectures. 

\item GoldenTransformer provides a structured and reproducible experimentation pipeline for analyzing the resilience of neural networks. The goal is not only to serve as a testing platform but also to aid in developing more robust LLM architectures.
\end{itemize}

\section{Related Work}

Fault injection has long been used in hardware and software testing to evaluate system robustness. ~\cite{9926241} demonstrates that single bit flips errors can cause minor application-level perturbations in DNNs. Research into neural network fault tolerance has expanded as deep learning systems become integral to mission-critical applications. Existing tools like TensorFI ~\cite{chen2020tensorfi, li2018tensorfi, narayanan2022fault}, PyTorchFI ~\cite{mahmoud2020pytorchfi}, and GoldenEye ~\cite{mahmoud2022goldeneye} have offered fault injection capabilities for CNNs and smaller-scale architectures. GoldenEye, in particular, introduced a flexible fault simulator for evaluating numerical formats and single- or multi-bit corruption in PyTorch-based DNNs. Similarly, BinFI ~\cite{10.1145/3295500.3356177} outlined a comprehensive framework to locate critical bits through fault injection, but limited their approach to CNN architectures only. These tools focus predominantly on CNNs or general feedforward architectures and often provide limited support for transformers. Moreover, their injection techniques tend to be coarse-grained, such as zeroing out random neurons or weights, which is insufficient for studying the structured and hierarchical nature of LLMs.

Most existing transformer robustness studies focus on adversarial inputs, corrupted training data, or model distillation. For example, TextAttack~\cite{DBLP:journals/corr/abs-2005-05909} established a framework for LLM adversarial attacks based on the mutation of input sequences to language models. While this and other approaches such as ~\cite{DBLP:journals/corr/abs-1803-01128}~\cite{DBLP:journals/corr/abs-1903-06620}~\cite{liu2024autodangeneratingstealthyjailbreak} are crucial, they do not simulate low-level internal or mathematical faults. Some recent works have explored quantization-induced errors or hardware-aware pruning, but these approaches remain disjoint from general-purpose fault injection research.

GoldenTransformer builds on the conceptual foundation laid by GoldenEye and PyTorchFI but introduces LLM-specific functionality. This includes layer-level manipulation of attention modules, adaptive dropout-based fault models, and attention mask corruption—capabilities not present in prior work. To our knowledge, this is the first transformer-specific open-source fault injection toolkit.

\section{System Design}
GoldenTransformer is built on PyTorch ~\cite{DBLP:journals/corr/abs-1912-01703} and integrates seamlessly with HuggingFace models and datasets. The core components are (1) the Fault Injector, which is the central interface for fault injection, (2) the Fault Modules, or BaseFault subclasses that implement different injection mechanisms, (3) the Experiment Runner, which coordinates injection, evaluation, and results logging with consistent configuration of faults, datasets, and performance metrics, (4) the Metrics class, including latency, accuracy, and perplexity, and (5) the visualization module, which converts experimental results into visual plots.
Each component is designed to be extensible, encouraging users to develop and test custom faults for different model architectures or tasks. Users can also design their own metrics to best suit their research goals.

\section{Fault Types}
GoldenTransformer supports a diverse set of fault models, including (1) layer-level faults such as attention mask corruptions or dropout faults, (2) activation-level faults such as clamp, noise, or random-zero corruption, (3) weight-level faults such as bit flips or random corruptions, and (4) attention-specific faults including mask noise corruptions or head dropout. Each type of fault exposes a severity parameter which controls the intensity or probability of the fault. For bit-flip faults, severity represents the probability of flipping a bit in the mantissa of a floating-point weight. For activation faults, the severity may control the magnitude of the noise or the fraction of activations affected. This allows researchers to systematically study the relationship between fault intensity and model degradation. Further faults may be designed and implemented by users as subclasses of the Base Fault. Thus, GoldenTransformer allows for completely customized experimental investigation of Transformer-based models through an easily extended base architecture.

\section{Demonstration and Use}
Fault injection is model-aware and adapts based on layer structure. For instance, LayerFault checks for layer lists named transformer.h or transformer, ensuring compatibility.
To ensure safe and reversible fault injection, the framework clones the original parameters when needed and supports fault rollback via a reversion method. Metric logging is handled by extensible classes such as Accuracy and LatencyMetric, which can be composed for multi-metric evaluation. Users may also design their own custom metrics in order to evaluate their model as best suits their needs.
Experiments can be configured with batch size, number of samples, and device target. Results are stored in timestamped directories with JSON logs, enabling reproducibility and easy tracking of experimental progress.

The code \ref{code:config} demonstrates a basic experiment setup using a HuggingFace model and the IMDB dataset. Users define the model, tokenizer, dataset, and list of faults to test. These are passed to the ExperimentRunner, which logs and stores the results in a timestamped directory. Output metrics include accuracy, latency, and confidence intervals.

\begin{mycode}
\begin{minted}[xleftmargin=18pt, linenos, frame=lines,framesep=1mm, baselinestretch=0.92, bgcolor=code_gray,fontsize=\footnotesize]{ini}
  faults = [
        LayerFault(layer_idx=0, severity=0.2),
        WeightCorruption(corruption_rate=0.1)
    ]
    metrics = [Accuracy(), LatencyMetric
    (num_runs=3)]
    runner = ExperimentRunner(
        injector=FaultInjector(model),
        faults=faults,
        metrics=metrics,
        dataset=dataset,
        batch_size=16,
        num_samples=50,
        device=device
    )
    results = runner.run()
\end{minted}
\captionof{listing}{GoldenTransformer Demonstration.}
\label{code:config}
\end{mycode}

\section{Experiments}
We conducted two experiments to evaluate the functionality and insight generation capabilities of GoldenTransformer. These experiments are designed as proof-of-concept evaluations and reflect preliminary results, with small-scale test samples used to enable rapid iteration and early understanding of transformer robustness profiles. While not definitive, the results demonstrate the framework’s flexibility across tasks and model architectures, as well as its capacity to surface meaningful differences in fault tolerance across layers. 
Furthermore, they demonstrate the nuanced robustness of transformers, offering insights that could guide future designs for fault-tolerant AI systems.
\subsection{Classification under Weight Faults (IMDB)}
\begin{figure}[!t]
    \centering
    \begin{minipage}{0.5\columnwidth}
        \centering
        \includegraphics[width=\linewidth]{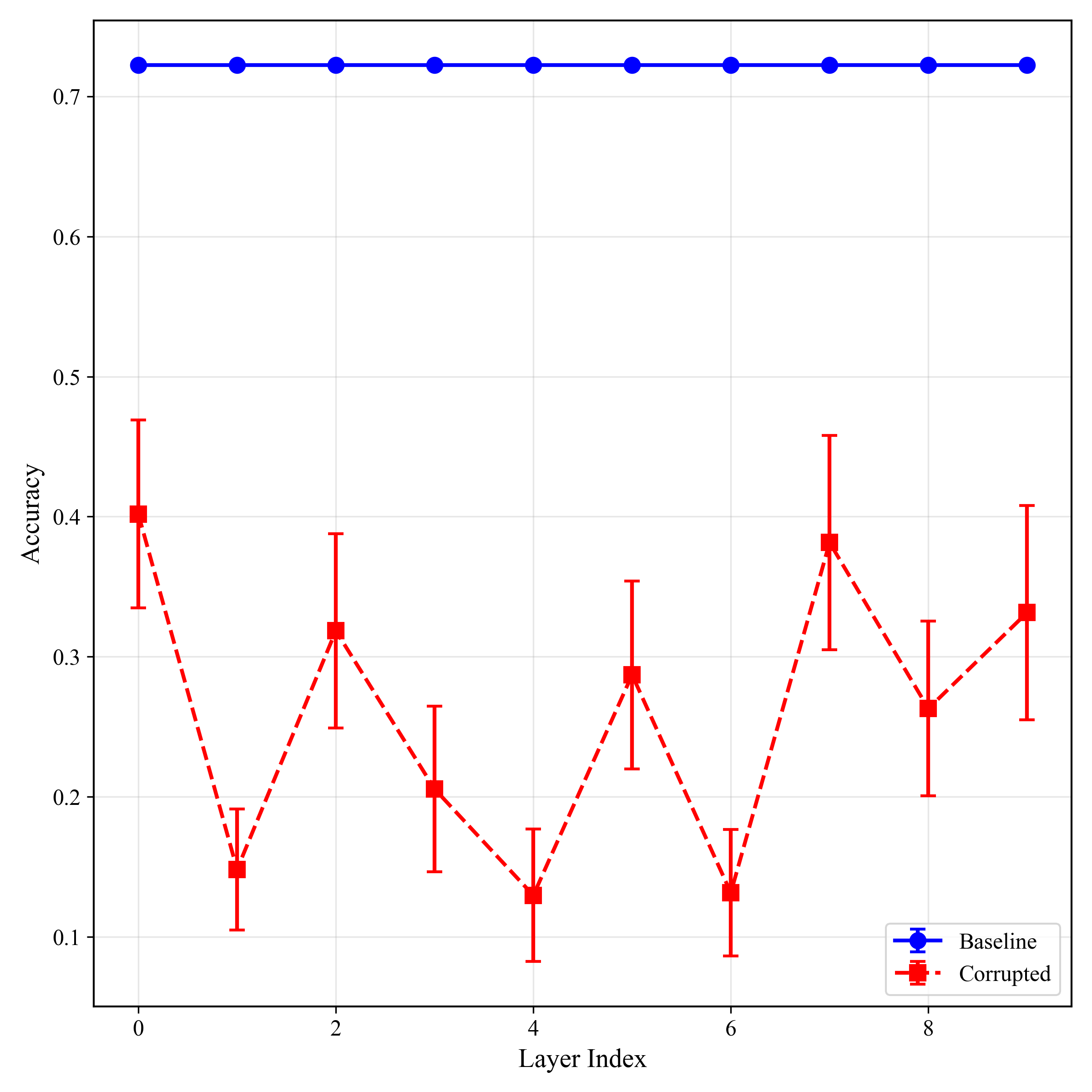}
    \end{minipage}%
    \hfill
    \begin{minipage}{0.5\columnwidth}
        \centering
        \includegraphics[width=\linewidth]{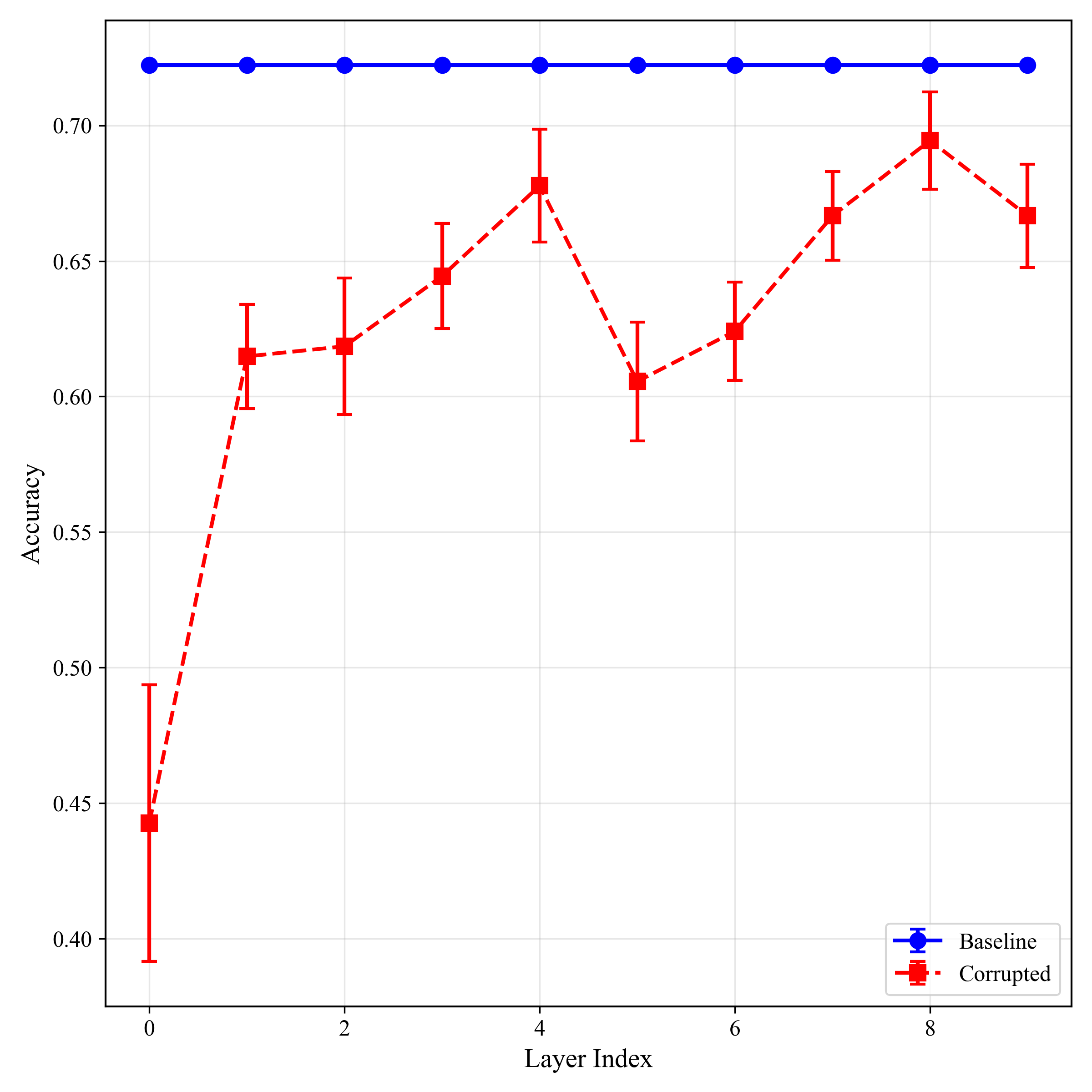}
    \end{minipage}
    \caption{DistilBERT on IMDB under Random Noise Injections with $p=0.1$ (left) and $p=0.05$ (right).}
    \label{fig:imdb-combined}
\end{figure}

To assess GoldenTransformer’s capacity to inject and track faults in classification models, we performed a series of weight corruption experiments on the textattack/distilbert-base-uncased-imdb model. This is a distilled version of BERT fine-tuned on the IMDB sentiment analysis dataset, allowing us to extend our framework beyond GPT-2 and demonstrate its applicability across encoder-only transformer architectures.
We injected random Gaussian noise into the weights of the model with a fixed corruption probability p, targeting each of the first 10 transformer layers individually. To control for variability in fault placement, we conducted 30 separate trials for each layer, varying the random seed from 42 to 71, and reported mean accuracy along with standard deviation to generate error bars. We used a randomly selected 50-sample subset from the IMDB test split for inference.
At $p=0.05$, random noise degraded the model’s accuracy across all layers, with average accuracy falling below the baseline in every case, as shown in Figure \ref{fig:imdb-combined}. All ten layers produced a statistically significant drop in accuracy, with non-overlapping error bars and a mean far below the baseline. Interestingly, layer 1 exhibited the lowest average accuracy but also the highest variance, suggesting early-layer corruption may lead to inconsistent but sometimes severe performance degradation. At $p=0.1$, the degradation effects were amplified and more consistent. All layers showed statistically significant drops in accuracy, with confidence intervals completely below the baseline and smaller standard error than layers in the previous experiment.
These experiments reinforce the notion of layer-dependent vulnerability, where some layers are more sensitive to fault injection than others—an insight that underscores GoldenTransformer’s capacity to reveal fine-grained structural robustness within transformer models. The results also highlight the complexity of fault tolerance in deep learning models and suggest that further, more systematic studies—potentially with larger datasets and repeated trials—are needed to fully characterize the relationship between fault severity and model robustness.

\subsection{Perplexity under Layerwise Bit Flips}
\begin{figure}
    \centering
    \includegraphics[width=0.5\linewidth]{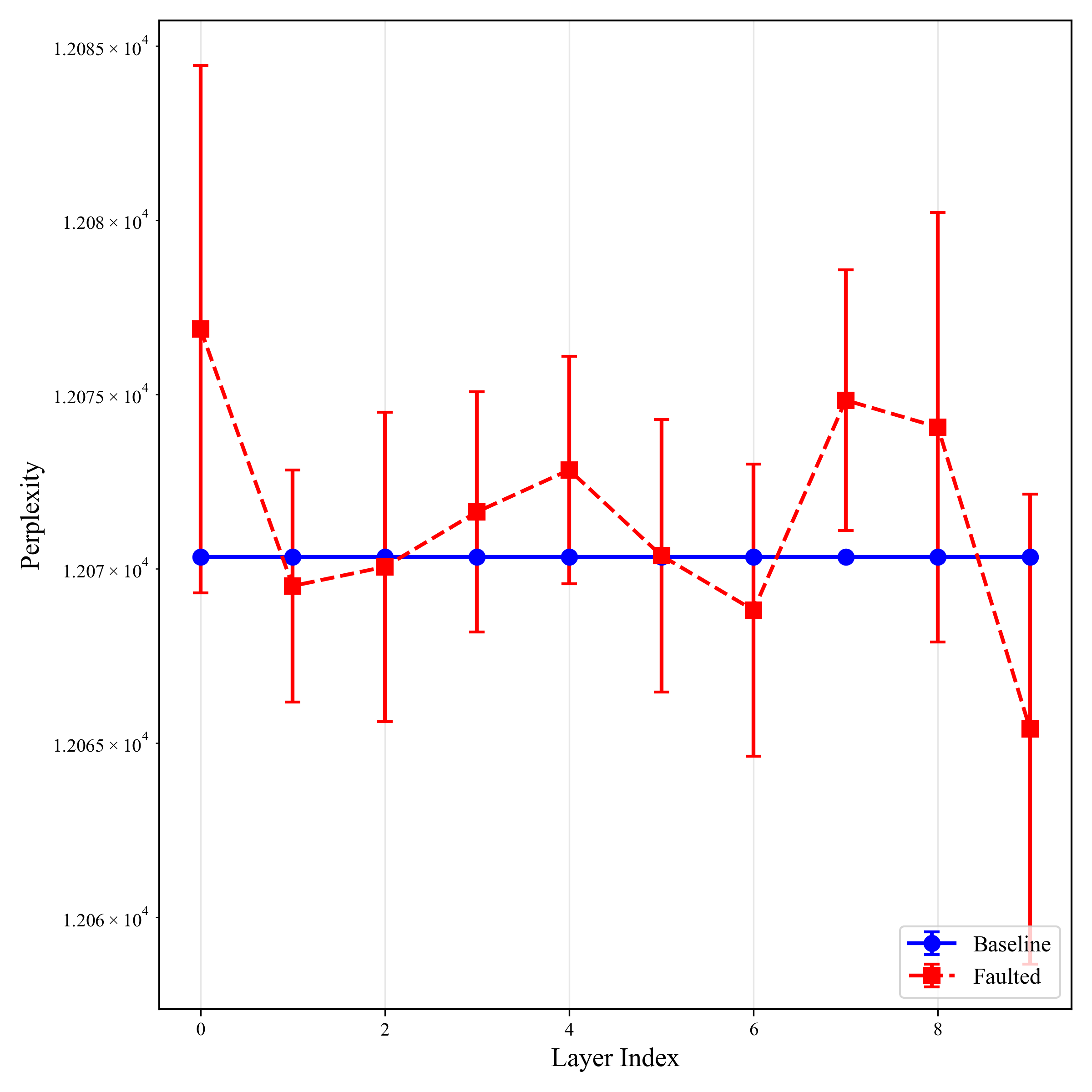}
    \caption{Perplexity under Layerwise Bit-Flip Faults for GPT2 on Wikitext2 }
    \label{fig:perplexity}
\end{figure}

We conducted a proof-of-concept fault injection experiment using the pretrained GPT-2 (117M) model from HuggingFace. To test sensitivity in a different modality and task, we applied GoldenTransformer to GPT-2 with greedy decoding (temperature 0) on the Wikitext2 language modeling benchmark, selecting the first 100 nonempty lines from the test split, truncating them to 32 tokens each, and injecting single-bit faults (restricted to mantissa bits) into the first 10 transformer layers, again using 30 seeds (42-71) for robustness. The evaluation metric was log-scale perplexity. As shown in Figure \ref{fig:perplexity}, only layer index 7 caused a statistically significant increase in perplexity— all other 95\% confidence intervals overlapped with the baseline. Although GPT2 appears resilient to low-level mantissa bit flips as there was variable impact on model perplexity, even small numerical perturbations can subtly affect its language modeling performance.  

\section{Discussion}
One of the primary implications is the differential sensitivity across layers, which challenges the assumption that transformer layers contribute equally or symmetrically to model behavior. Faults in early layers can, in certain settings, cause disproportionate degradation—pointing to critical paths in information propagation. This insight can inform a wide array of engineering decisions: from robustness-aware training (e.g., layer-specific regularization), to more efficient deployment strategies (e.g., selectively hardening vulnerable layers in edge deployments).

That said, the current methodology has limitations. The Gaussian noise model is synthetic and does not fully capture the spectrum of real-world bit-level faults, and the experiments assume a relatively static architecture. Future work could expand this to dynamic tasks, real hardware, and alternative corruption patterns. 

\section{Conclusion}
GoldenTransformer is a modular, extensible, and LLM-specific fault injection framework that provides researchers and engineers with tools to evaluate and improve the resilience of transformer-based models. We injected random Gaussian noise into the weights of a finetuned DistilBERT model and evaluated on the IMDB sentiment classification dataset, observing significant degradation in classification accuracy across all layers. However, some layers observed more variability and more significant drops in accuracy, suggesting nonuniform layer sensitivity to fault injections.
We also conducted layerwise mantissa bit flips on a pretrained GPT2 117M model applied on the Wikitext2 benchmark, observing resulting changes in log-scale model perplexity, and did not see significant impact. This result indicates that more systematic exploration is required to observe the effect of bit-flip faults in generative LLMs.
Importantly, GoldenTransformer is not merely a diagnostic tool; it opens a potential design loop. Its findings could guide targeted model compression or fault-aware finetuning. Moreover, from a systems perspective, the framework could be adapted to quantify reliability under hardware noise or adversarial conditions, especially for safety-critical or resource-constrained environments.

\section*{Acknowledgment}
The author would like to thank Dr. Xun Jiao and Mr. Jinghao Wen at Villanova University for their mentorship and advice in this project.

\bibliographystyle{IEEEtran}
\bibliography{refer}

\end{document}